\documentclass[conference]{IEEEtran}
\IEEEoverridecommandlockouts
\usepackage{cite}
\usepackage{amsmath,amssymb,amsfonts}
\usepackage{algorithmic}
\usepackage{graphicx}
\usepackage{textcomp}
\usepackage{color,array}
\usepackage{graphicx}
\usepackage{algorithm}
\usepackage{algorithmic}
\makeatother
\usepackage[utf8]{inputenc}
\usepackage[table]{xcolor}
\usepackage{tabularx,booktabs}
\newcolumntype{C}{>{\centering\arraybackslash}X} 
\usepackage{cite}
\usepackage{amsmath}
\usepackage{amsmath,amssymb,amsfonts}
\usepackage{algorithmic}
\usepackage{graphicx}
\usepackage{textcomp}
\usepackage{xcolor}
\usepackage{hyperref}
\usepackage{caption}
\usepackage{pifont}
\usepackage{xcolor}
\usepackage{rotating}
\usepackage{rotfloat}
\usepackage{algorithm}
\usepackage{algorithmic}
\usepackage{gensymb}
\usepackage{textcomp}
\usepackage{placeins}
\usepackage{balance}
\usepackage{subfigure}
\usepackage{booktabs}
\usepackage{tabularx}
\usepackage{array}
\usepackage{pbox}
\usepackage{longtable}
\usepackage{booktabs}

\def\BibTeX{{\rm B\kern-.05em{\sc i\kern-.025em b}\kern-.08em
    T\kern-.1667em\lower.7ex\hbox{E}\kern-.125emX}}
\begin{document}

\makeatletter
\newcommand{\newlineauthors}{%
  \end{@IEEEauthorhalign}\hfill\mbox{}\par
  \mbox{}\hfill\begin{@IEEEauthorhalign}
}
\makeatother

\title{Video Vision Transformers for Violence Detection}

\author{\IEEEauthorblockN{Sanskar Singh}
\IEEEauthorblockA{\textit{IIIT Naya Raipur}\\
sanskar21102@iiitnr.edu.in}
\and
\IEEEauthorblockN{Shivaibhav Dewangan}
\IEEEauthorblockA{\textit{IIIT Naya Raipur}\\
shivaibhav21102@iiitnr.edu.in}
\and
\IEEEauthorblockN{Ghanta Sai Krishna}
\IEEEauthorblockA{\textit{IIIT Naya Raipur}\\
ghanta20102@iiitnr.edu.in}

\newlineauthors
\IEEEauthorblockN{Vandit Tyagi}
\IEEEauthorblockA{\textit{IIIT Naya Raipur}\\
vandit21102@iiitnr.edu.in}
\and
\IEEEauthorblockN{Sainath Reddy}
\IEEEauthorblockA{\textit{IIIT Naya Raipur}\\
sankepally20102@iiitnr.edu.in}
\and
\IEEEauthorblockN{Prathistith Raj Medi}
\IEEEauthorblockA{\textit{IIIT Naya Raipur}\\
prathistith19102@iiitnr.edu.in}
}

\maketitle

\begin{abstract}
Law enforcement and city safety are significantly impacted by detecting violent incidents in surveillance systems. Although modern (smart) cameras are widely available and affordable, such technological solutions are impotent in most instances. Furthermore, personnel monitoring CCTV recordings frequently show a belated reaction, resulting in the potential cause of catastrophe to people and property. Thus automated detection of violence for swift actions is very crucial. The proposed solution uses a novel end-to-end deep learning-based video vision transformer (ViViT) that can proficiently discern fights, hostile movements, and violent events in video sequences. The study presents utilizing a data augmentation strategy to overcome the downside of weaker inductive biasness while training vision transformers on a smaller training datasets.  The evaluated results can be subsequently sent to local concerned authority, and the captured video can be analyzed. In comparison to state-of-the-art (SOTA) approaches the proposed method achieved auspicious performance on some of the challenging benchmark datasets.
\end{abstract}

\begin{IEEEkeywords}
Violence Detection, Video Classification, Video Vision Transformers, Augmentation
\end{IEEEkeywords}

\section{Introduction}
Violent behavior in public places is becoming a serious threat to personal security and social order. The increase in violent behaviours in public areas can be ascribed to a number of variables. Individual greed, frustration, anger, and social and economic insecurity are the fundamental drivers of the surge in violence. Despite the fact that modern technology has boosted the capability of surveillance systems, there is a widespread upsurge in violence-related issues. According to the Global Burden of Disease research \cite{owidhomicides}, around 415,000 people died as a result of homicide in 2019 alone. This was roughly three times the number of people killed in armed conflict and terrorism. Another survey conducted by FBI \cite{2} , it was estimated that in the US only, there were 366.7 violent offenses per 100,000 denizens in the year 2019 that, which accounts for an approximate total of 1,203,808 crimes in the whole country. 

Ensuing the vogue of data mining methods in law enforcement, automatic violence detection has the competence to provide a quick response in the event of violence, minimizing delays in asking for aid when it may be a matter of life and death. The potential to perceive individuals' aggression in video footage would be extremely beneficial in real-time camera systems and motion picture data analysis. Real-time video systems could help to identify violence in situations where peace are required, for instance, in aircraft cabins, recreation centers, and educational institutes. Movie analyzing systems could incorporate automatic violence detection to rate movies while preventing youngsters from watching violent sequences. Consequently, an autonomous system considerably also minimizes the stress placed on a person who is expected to watch hours of footage. Therefore as a corollary, the computer vision community is becoming interested in this area of study. 

Serrano et al. \cite{8375994} created a dynamic strategy that combines a two-dimensional Convolutional neural network (2D-CNN) with Hough Forest (HF), with crucial information from HF utilized to represent an image in a sequence. An analogous approach for violence detection based on a 3D - CNN model was presented by Ullah et al. \cite{19112472}. The authors classified the violent scenes based on the spatial and temporal attributes of 3D CNN. Abdali et al. \cite{8852616} presented their work, which included CNN as a spatial feature collector and Long Short-Term Memory (LSTM) as a temporal connection learning approach.

Contrary to the CNN-based approach in computer vision, Dosovitskiy et al. \cite{DBLP:journals/corr/abs-2010-11929} proposed a pure transformer-based framework, Vision Transformer (ViT), which, when applied directly to image patch sequences, can perform very well on image classification tasks. ViT achieves excellent results compared to state-of-the-art convolutional networks when pretrained on vast amounts of data and transferred to several mid-sized or small picture recognition benchmarks. ViT also ensures the usage of significantly fewer computation resources to train. Arnab et al. \cite{DBLP:journals/corr/abs-2103-15691} presented a pure Transformer-based approach for video classification, Video Vision Transformer (ViViT). ViViT is an extension of vision transformer, considering the fact that each video consists of several image frames. The fundamental computation performed in this architecture is self-attention which can be described as concentrating one's attention on oneself. Furthermore, ViT can be utilized for real time deployment due to it's less computational efforts \cite{https://doi.org/10.48550/arxiv.2209.01401}. The current study implements the Spatio-temporal attention variant of the video vision transformer (ViViT) for the violence classification task. It aggregates Spatio-temporal tokens from the input video and encodes them using a network of transformer layers. 

\section{Related Works}

Various approaches for violence detection based on handcrafted features and deep features have been developed. Following section gives a brief overview of the past works that have been presented before based on the both the features.

\subsection{Handcrafted Features-Based Approaches}

Datta et al. \cite{1044748} employed the trajectory of motion information and limb orientation of a person in the scene to detect violence. 
According to a study provided by Nguyen et al. \cite{1467545}, the hierarchical hidden Markov model (HHMM) might be used to identify aggressive behaviour. Their primary contribution is the application of a common HHMM framework for the detection of violence.
Kim and Grauman \cite{5206569} use a mixture of probabilistic PCA models to simulate local optical flow patterns and a Markov Random Field (MRF) to guarantee global consistency. On contrast to the above suggested methods, Mahadevan et al. \cite{5539872} suggests that the representations based on optical flow are not robust enough to detect anomalous occurrences in terms of joint appearance and motion. Mahadevan et al. devised a method for recognising violent scenarios by continuously monitoring blood and flames as well as the degree of motion and loudness. Nievas, E.B. et al. \cite{10.5555/2044575.2044624} introduced a novel bag-of-words (BoW) architecture for movement recognition in the context of conflict detection, which makes use of descripting action techniques like motion scale-invariant feature transform (MoSIFT) and space-time interest points (STIP).

\subsection{Deep Learning-Based Approaches}

The recent advancement of deep neural networks in activity recognition has inspired numerous studies to highlight the application of neural networks in violence detection task. Dong et al. \cite{10.1007/978-981-10-3002-4_43} recommended using three streaming deep neural networks to gain multiple forms of violent information from raw videos, namely spatial, temporal, and acceleration streams.
Fenil et al. \cite{SAMUELR2019191} have published a violent action identification system for a soccer game. by deriving a histogram of oriented gradient (HoG) characteristics from each frame which were utilised to train and ensure the use of bidirectional long short-term memory (BD-LSTM) for both forward and backward information access. Sudhakaran et al. \cite{8078468} devised a convolutional neural network in conjunction with a convolutional-long short term memory (ConvLSTM) to be capable of recording localised spatio-temporal data, allowing for the analysis of localized motion in video.  Ullah et al. \cite{19112472} improved the 3-dimensional convolution neural network (3D CNN) model by transforming the training model to an intermediate representation and using open visual inference and neural networks to detect violent events autonomously. However, a network cannot learn long temporal information with such an architecture. Similar to this, Accattoli et al. \cite{doi:10.1080/08839514.2020.1723876} employed a 3D CNN architecture to capture motion data without background information, that can then be used as an feed for a SVM (linear) to categorise video sequences as violent or peaceful.

\section{Proposed Methodology}

The following section discusses the proposed novel framework. This section is further composed of 3 subsections to get along the various phase of the work. Pre-processing part gives the various analogies applied to the video such as conversion into frames and resizing such frames. Once the videos are converted into several frames, the converted frames are applied few image augmentation techniques for effective training of the model and is discussed under the Video Frames Augmentation part. The last subsection deals with the implementation of Video Vision Transformer (ViViT)  on the retrieved augmented frames. Videos are then classified into violent or non-violent using the ViViT framework as discussed later. The study has shown significant results on some of the challenging benchmark datasets which are widely acknowledged for violence detection.
\begin{figure}[ht]
    \centering
    \includegraphics[width=\linewidth,height=37mm]{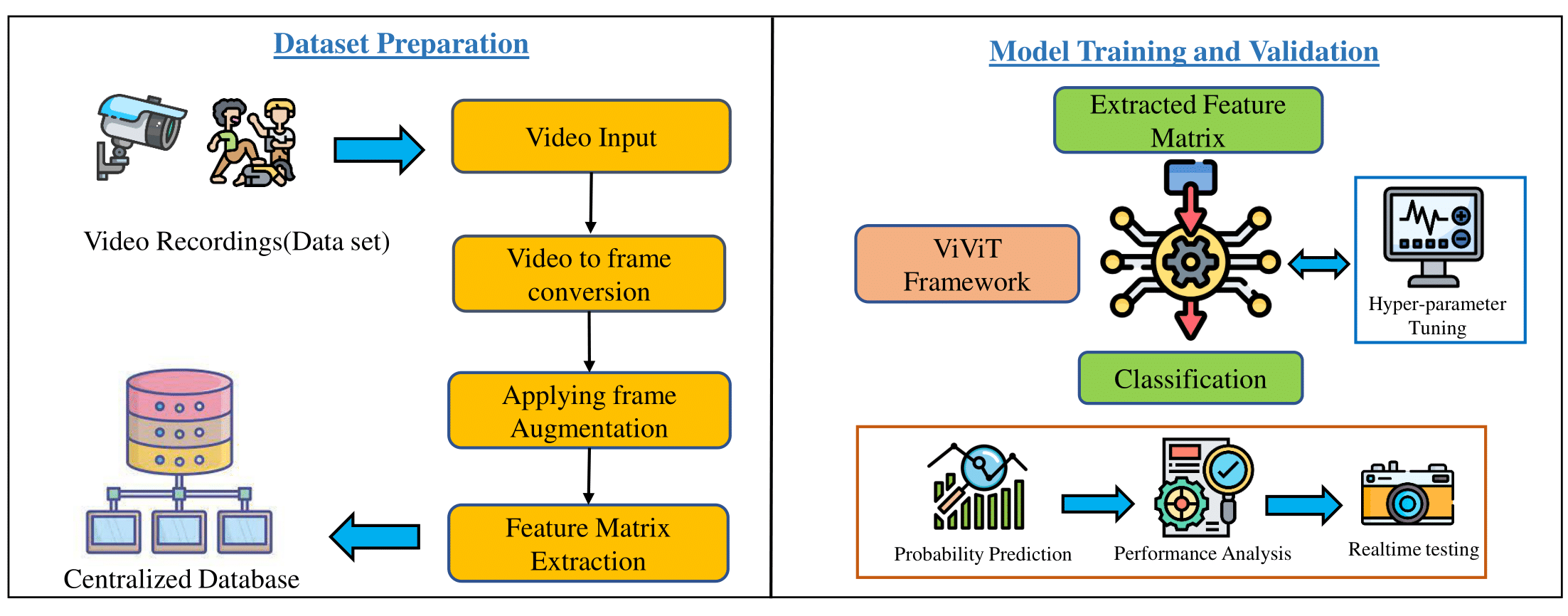}
    \caption{Overview of the framework}
    \label{img1}
\end{figure} 

\subsection{Pre-processing}

At first, video data extracted from various datasets are converted into several frames of fixed counts. Feeding frames with higher pixels to the model can lead to intensive computations. To avoid this, the frames are scaled down to a smaller resolutions. Also, the aspect ratio must remain the same while scaling down, because change of aspect ratio may lead to losing some of the crucial information in the video. The work is done with the premise that 56 consecutive frames are adequate to illustrate a violent event because the usual frame rate produced by a video clip is around 25-30 frames per second. Hence, each video is divided into a frames of 56 and further, each frame is resized to smaller fixed pixel sizes while considering the concept of same aspect ratio between the height and width channels.

\subsection{Video Frames Augmentation}
Due to Transformers lacking some of convolutional networks' inductive biases,ViT has only been demonstrated to be effective if pre-trained on large-scale datasets. Taking inspirations from the study presented by Steiner et al. \cite{DBLP:journals/corr/abs-2106-10270}, the present work judiciously implies data augmentation technique to train much better models on a dataset of a given size. Thus, the pre-processed video frames are then transformed with several image augmentation techniques such as Gaussian Blur, Random rotation, uniform perturbations and flipping in various directions. Figure 2 demonstrates the various transformations applied to a sample video frame. 

\begin{figure}[h]
    \centering
    \includegraphics[width=\linewidth]{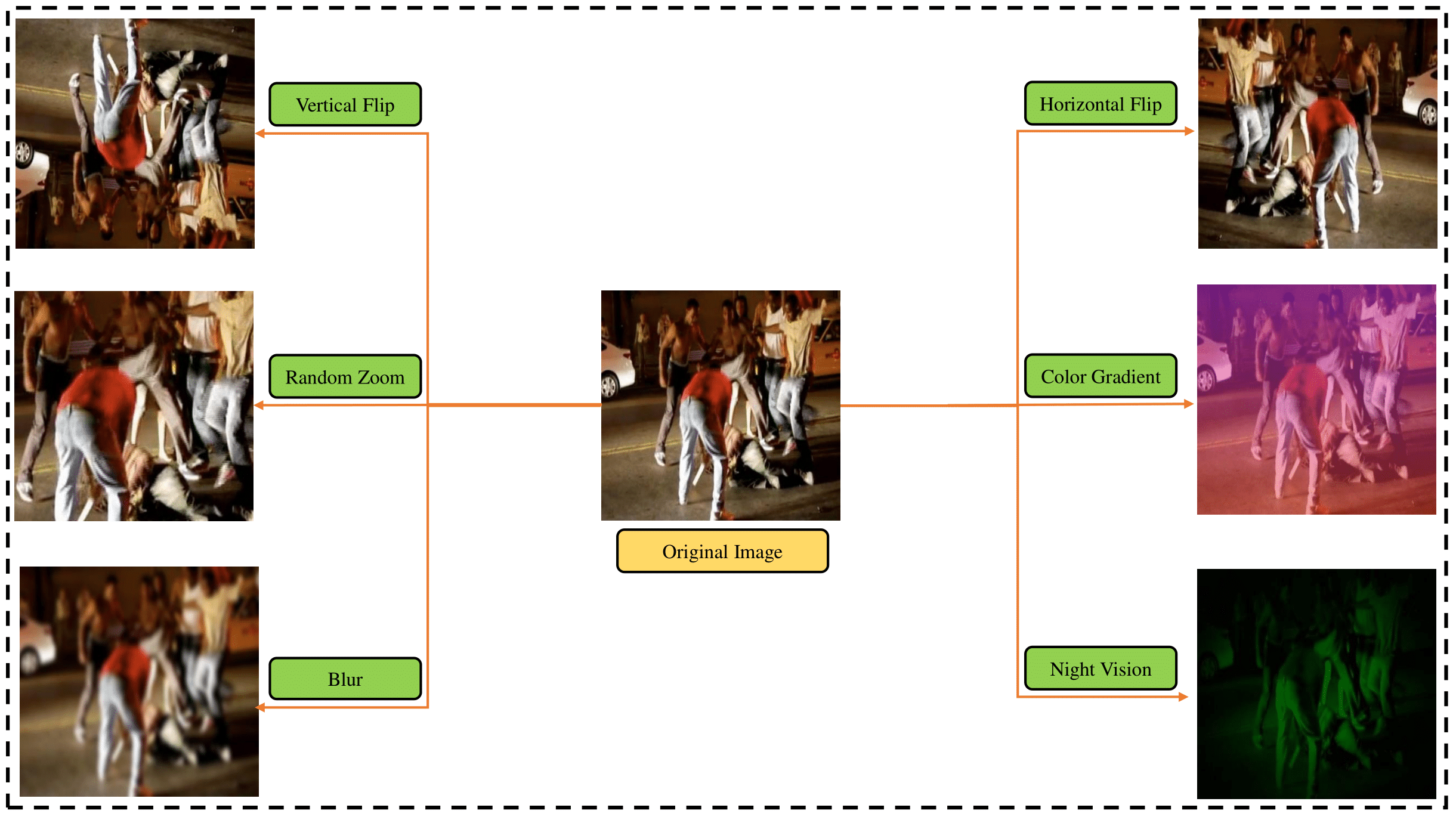}
    \caption{Single Frame Augmentation}
    \label{img2}
\end{figure}

\subsection{Video Vision Transformers for Video Classification}
After data preprocessing and augmentation, an efficient machine learning architecture is required to fulfil video classification task. The current research methodology uses video vision transformers (ViViT) for effective video classification. 
\subsubsection{Embedding}
Given a video clip $Z \in \mathbb{R}^{T \times W \times H \times C}$, where $T$ defines the duration of the clip, $W$ and $H$ signify the width and height of the single frame, and $C$ specifies the number of channels, the video clip $Z$ is converted to a series of tokens $\tilde{\mathbf{y}} \in$ $\mathbb{R}^{n_{t} \times n_{w} \times n_{h} \times d}$.
To this, the proposed ViViT framework employs tubelet embedding method  for mapping a video to a sequence of tokens. 

\begin{figure}[h]
    \centering
    \includegraphics[width=0.8\linewidth]{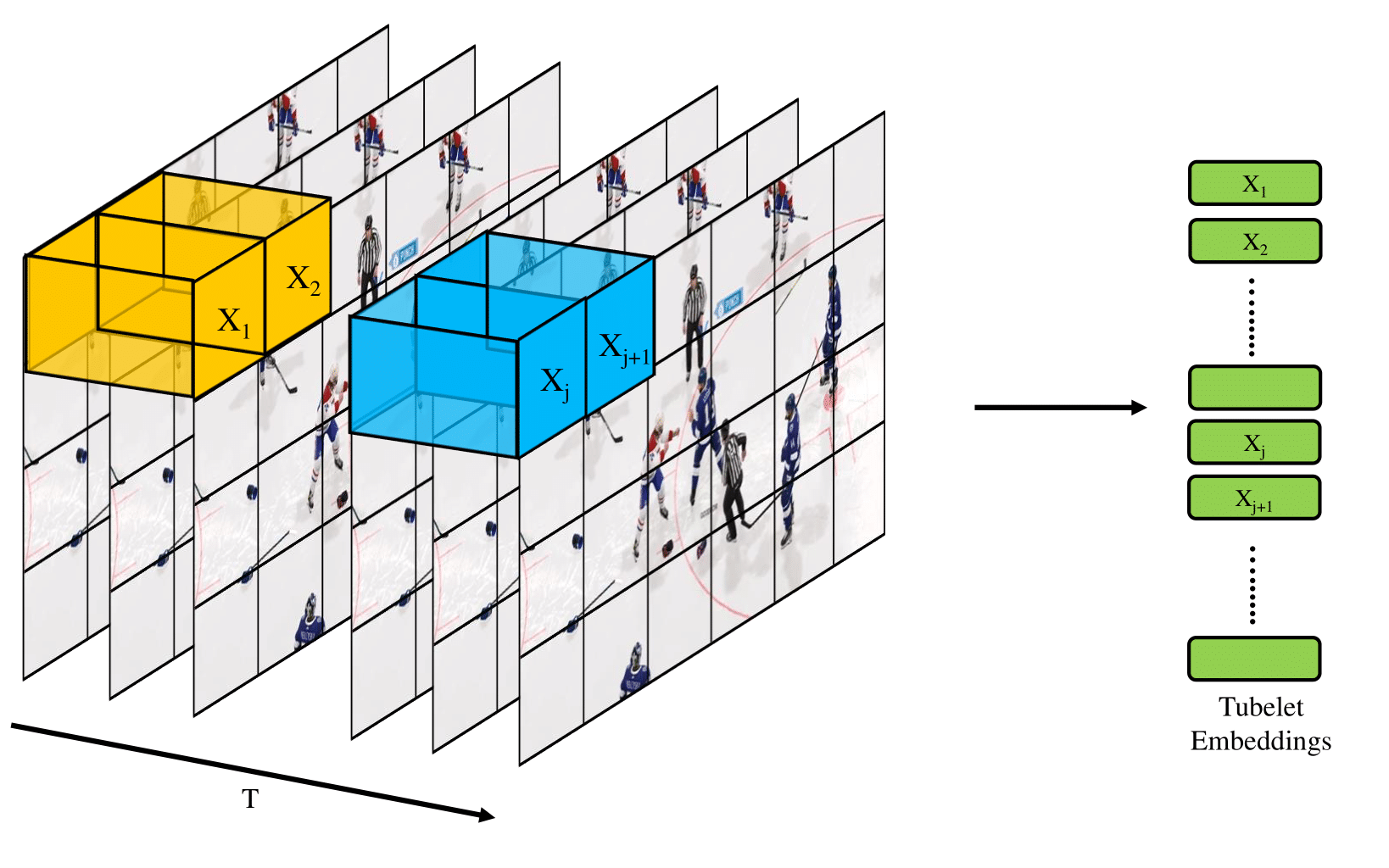}
    \caption{Tubelet Embedding}
    \label{img3}
\end{figure}

So, Instead of extracting patches from each frame as proposed in ViT's, non-overlapping, spatiotemporal tubes of the input sequences are retrieved from the video frames. These volumes of sequences include both temporal as well as frame-specific patches.
The obtained volume of patches are then linearly flattened to build several video encoded tokens. For a tubelet with dimension $t \times w \times h $,
\begin{equation}
   n_{t}=\left\lfloor\frac{T}{t}\right\rfloor, \hspace{0.5em}
   n_{w}=\left\lfloor\frac{W}{w}\right\rfloor
   \hspace{0.5em} and \hspace{0.5em}      
   n_{h}=\left\lfloor\frac{H}{h}\right\rfloor
\end{equation}

encoded tokens are retrieved from the temporal, width and height dimensions respectively.Greater numbers of tokens are implied by tubelet with smaller dimensions, which further improves computation. This method, intuitively, integrates spatio-temporal information during tokenisation. Drawing cues from the original BERT study, an additional CLS token ($y_{cls}$) is added to the set of embedded tokens as shown in Figure 4(a), which is responsible for aggregating global video frame information and final classification. Since the transformer's succeeding self-attention procedures are permutation invariant, a learned positional embedding, $P \in \mathbb{R}^{N \times d}$, is also appended to the tokens to maintain track of their positional information.

The present study suggest the use of \textbf{Spatio-Temporal Attention} mechanism of ViViT architecture, proposed by Arnab et. al., for violence classification task. Hence, after tokenizing videos with a Tubelet embedding-based technique, all spatio-temporal tokens retrieved are passed on directly through a standard transformer encoder. The sequence
of tokens input to the following transformer encoder is 
\begin{equation}
    \textbf{y}\hspace{.2em}=\hspace{.2em}[y_{cls},\textbf{J}x_1,
    \textbf{J}x_2,...,\textbf{J}x_N]\hspace{.2em}+\hspace{.2em}P 
\end{equation}
where \textbf{J} denotes to the patch embeddings.

\subsubsection{Transformer Encoder} The architecture of the Transformer Encoder is made up of several stacks of $L$ identical blocks. Each block begins with a Multi-Head Self Attention (MSA) layer and ends with a Multi-Layer Perceptron (MLP) blocks.  As illustrated in the mathematical formulae (3) and (4), both sub-components of the transformer encoder operate with a normalisation layer (LN) followed by residual skip connections.The model's sub-layers and embedding layers all create an output of embedded dimension $D$. The preceding step's $y$ vector is transferred through the transformer encoder architecture to produce the context vector $C$. 
\begin{equation}
    Y_l\hspace{.1em}=\hspace{.1em}y_{l-1} + \hspace{.1em}MSA(LN(y_{l-1}))\hspace{.1em} \hspace{2em}l=1,...,L
\end{equation}
\begin{equation}
    C_{(l+1)}\hspace{.1em}= \hspace{.1em}Y_l + \hspace{.1em}MLP(LN(Y_l))\hspace{.1em} \hspace{1.5em}l=1,...,L
\end{equation}

\begin{figure*}[ht]
    \centering
    \centerline{\includegraphics[height=98mm,width=175mm]{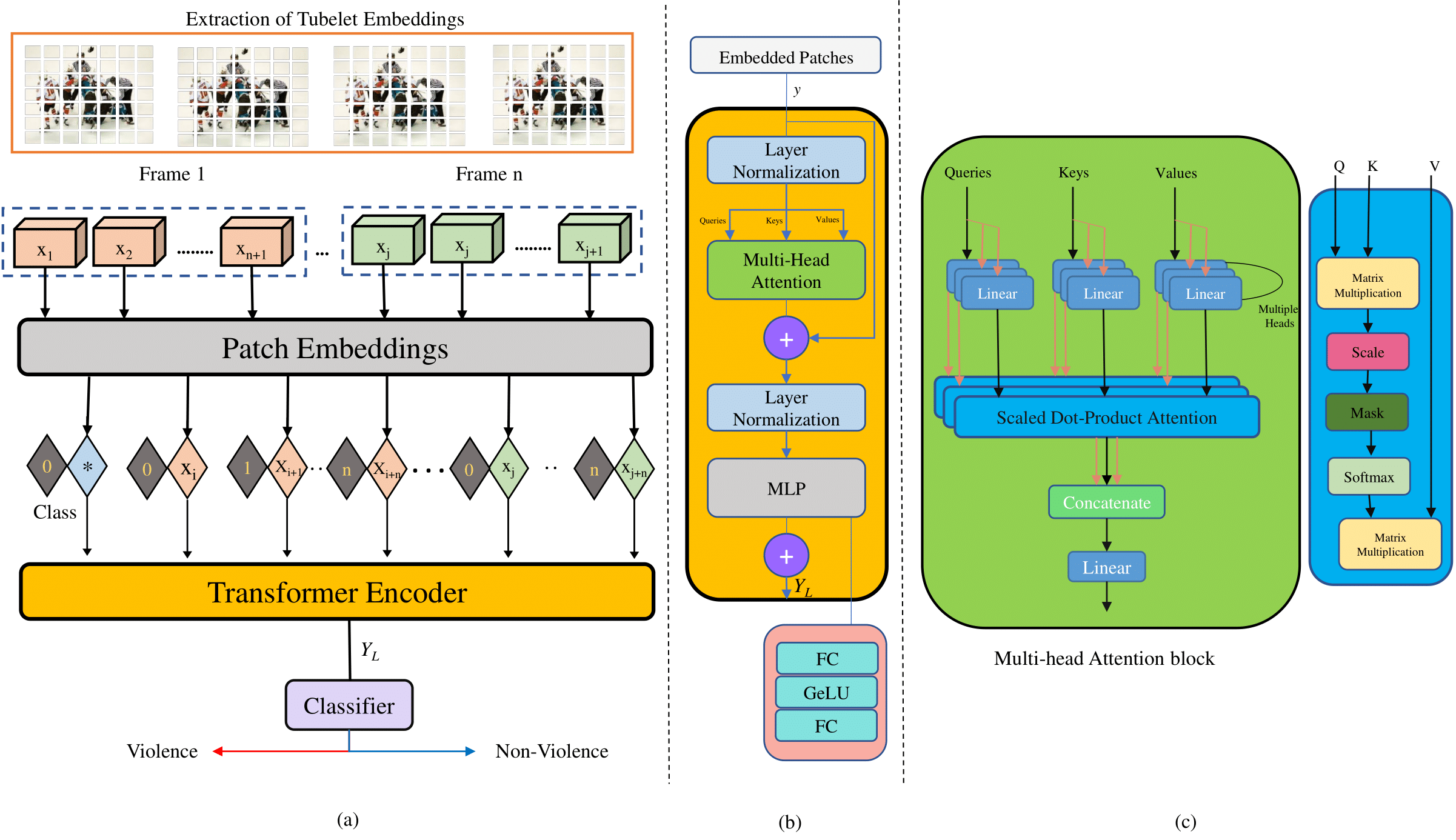}}
    \caption{(a) Generation of patch embeddings and Conceptual overview of ViViT model (b) Transformer Encoder (c) Multi head Attention block}
    \label{img_4}
\end{figure*} 

Fig. 4 (c) provides an illustration of the MSA block's procedures. The Scaled Dot-Product Attention or self-attention (SA) is the primary backbone of a Multi-Head Attention unit. SA enables the Transformer to discover meaningful relationships between input tokens. To begin, the input $y$ vector is transferred into three new matrices for each SA component: Q (query), K (key), and V (value) by multiplication with learnable weight matrices: $W_q$, $W_k$, and $W_v$, respectively.
\begin{equation}
     y \hspace{.3em} \times\hspace{.3em} W_q = Q 
\end{equation}
\begin{equation}
     y\hspace{.3em} \times \hspace{.3em}W_k = K
\end{equation}     
\begin{equation}
     y\hspace{.3em} \times \hspace{.3em}W_v = V 
\end{equation}
The Queries $Q$ are then multiplied by the transpose of Keys $K^T$, and the obtained result vector $Y$ is divided by the square root of the dimension $D$ to overcome the vanishing gradient problem.
This matrix is then sent through a Softmax activation layer and multiplied by the Values $V$ to produce the final output known as Head $H$. 
\begin{equation}
    Q \hspace{.3em} \times \hspace{.3em}K^T= \frac{Y}{\sqrt{D}} 
\end{equation}
\begin{equation}
   H\hspace{.3em}=\hspace{.3em} Softmax\hspace{.3em}(\frac{Y}{\sqrt{D}})  \hspace{.3em} \times \hspace{.3em}V= W_{attention} \hspace{.3em} \times \hspace{.3em} V 
\end{equation}
The Scaled Dot-Product Attention is applied $h$ times (h=8) to obtain $h$ attention heads. Thus a total of 8 attention heads are applied. Concatenating the results of each attention head, a feed-forward layer with learnable weights $W_0$ is then applied, as demonstrated in (10).
\begin{equation}
   MSA\hspace{.3em}=\hspace{.3em} concat(SA_1,SA_2,...,SA_N) \hspace{.3em} \times \hspace{.3em} W^0
\end{equation}
\begin{equation}
    \textbf{C}\hspace{.2em}=\hspace{.2em}[c_0,\textbf c_1,
    \textbf c_2,...,\textbf c_N]\hspace{.2em} 
\end{equation}

The MLP block structure is composed up of feed-forward dense layers that are fully coupled and have GeLU non-linearity.
The output of the encoder block is the context vector $C$, given in (11). Once context vector $C$ is collected, just context token, $c_0$ is required for classification. This context token $c_0$ is processed through an MLP (Multi-Perceptron Layer) head and softmax activation to yield a probability distribution of the target label for the video. The MLP head is implemented in the pre-training stage with one hidden layer and $tanh$ as non-linearity, and in the fine-tuning step with a single linear layer. Based on the probability distribution of the target label, the video is classified into violent or non-violent.

\section{Experimental Results}
The evaluation of performance of the model has been done by taking into consideration of the two challenging benchmark datasets in the domain of violence detection. In order to conduct the experiments with the maximum precision, fine-tuning of  several hyperparameters in the model such as learning rates, patch size and weights have been done. A report depicting classification accuracy and F1 score has also been presented. The redeemed accuracy are further compared with the results from previously proposed  state-of-the-art methods. 
Table 1 includes a brief representations of the datasets used for the study.
\begin{table}[ht]

\caption{Statistical Description of the datasets }
\label{table1}
\begin{center}
\scalebox{.9}
{
\begin{tabular}{|c|c|c|c|c|}
\hline
\textbf{Datasets} & \textbf{Samples Present} & \textbf{Resolution} & \textbf{Violent Clips} & \textbf{Non-Violent Clips}\\ 
\hline
Hockey Fight & $1000$ & $360 \times 288$ & $500$ & $500$\\
\hline
Violent Crowd & $246$ & $320 \times 240$ & $123$ & $123$\\
\hline
\end{tabular}
}
\end{center}
\end{table}

\subsection{Evaluation of the ViViT model}

The evaluation of the presented model in the pertinent literature of violence detection is discussed in the subsequent section. 
The performance of the model for crowd and hockey fight violence dataset was increased by including three tasks into the training of the effective ViViT model at once: computing output, troubleshooting faults, and fine-tuning hyper-parameters. The model achieves the optimum training and validation accuracies with a 60-40 split in both datasets. The precise set of hyper-parameters that produced the highest degree of accuracy after several tuning iterations are displayed in Table III.

\begin{figure*}[ht]
\centerline{\includegraphics[height=40mm,width=148mm]{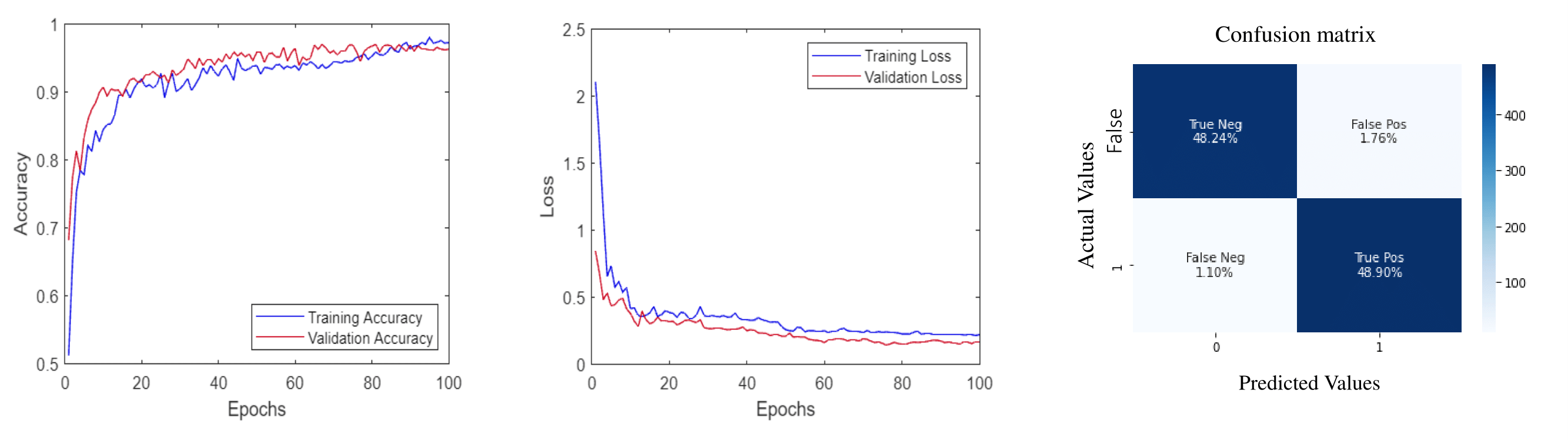}}
\caption{Learning curves of training and validation accuracy and loss for hockey fight and its confusion matrix}
\label{fig7}
\end{figure*}

\begin{figure*}[ht]
\centerline{\includegraphics[height=40mm,width=148mm]{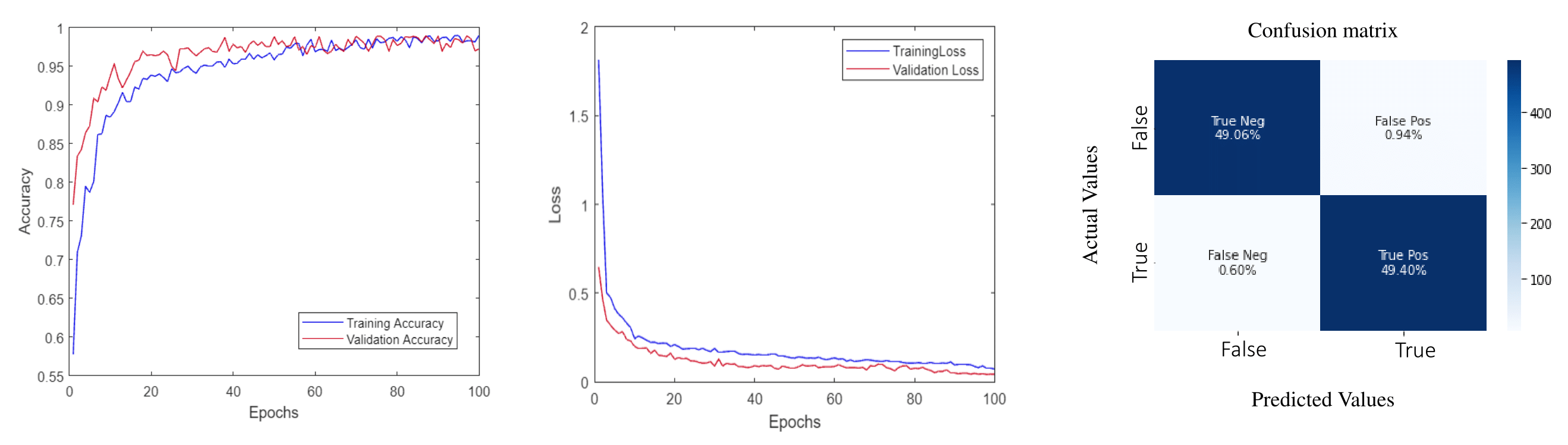}}
\caption{Learning curves of training and validation accuracy and loss for crowd violence and its confusion matrix}
\label{fig8}
\end{figure*}

\begin{table*}[t]
	\centering
	\caption{Computed Evaluation Metrics}
	\label{table7}
	\begin{tabular}{|c|c|c|c|c|c|c|c|c|}
		\hline
		
		\multicolumn{1}{|c}{\textbf{}} &
		\multicolumn{4}{|c|}{\textbf{Hockey Fight}}  &
		\multicolumn{4}{|c|}{\textbf{Violent Crowd}}  
		\\
		
		\hline
		\textbf{} &
		Precision&
		Recall &
		F1 score &
		 Video Support &
		 Precision&
		Recall &
		F1 score &
		 Video Support 
			\tabularnewline
		\hline

		 Violence  & 
	     0.98 &
	     0.98 &
	     0.98 &
	     196 &
	     0.99 &
	     0.99 &
	     0.99 &
	     49
		\tabularnewline
		\hline
		
		 Non - Violence  & 
	     0.97 &
	     0.97 &
	     0.97 &
	     194 &
	     0.98 &
	     0.97 &
	     0.97 &
	     47 
		\tabularnewline
		\hline
		
		 Macro Average  & 
	     0.97 &
	     0.97 &
	     0.97 &
	     390 &
	     0.98 &
	     0.98 &
	     0.98 &
	     96 
		\tabularnewline
		\hline
		
		Weighted Average  & 
	     0.97 &
	     0.97 &
	     0.97 &
	     390 &
	     0.98 &
	     0.98 &
	     0.98 &
	     96
		\tabularnewline
		\hline

	\end{tabular}
\end{table*}

\begin{table}[ht]
\caption{Tuned Hyper-parameters for the model}
\label{table2}
\begin{center}
\scalebox{1.1}
{
\begin{tabular}{|c|c|}
\hline
\textbf{Hyper-parameter} & \textbf{Attribute}  \\ 
\hline
Number of classes & 2 \\
\hline
Batch Size & 32\\
\hline
Patch Size & (8,8,8)\\
\hline
Epochs & 100\\
\hline
Learning Rate & 0.0001\\
\hline
Weight Decay & 0.00001\\
\hline
Projection Dimension & 128\\
\hline
Number of Heads & 8\\
\hline
Number of Layers & 8\\

\hline
\end{tabular}
}
\end{center}
\end{table}

 After multiple iterations of fine-tuning the hyperparameters, the model trained on hockey fight dataset yielded a maximum training accuracy of 96.57\% and validation accuracy of 97.14\%. Similarly the model trained on crowd fight violence dataset, the maximum training and validation accuracy achieved are 98.73 \%  and 98.46 \% respectively. The learning curves of accuracy and loss for training and validation of the models for both hockey fight and crowd violence dataset obtained are shown in Fig. 5 and Fig. 6 respectively. These learning plots explicitly show a well-fitted learning algorithm since both the validation and training curves retain a stable point with slightest gap. 

Precision measures such as Precision, Recall, and F1 scores were also estimated and portrayed for further assessment of the model. These pertain to obtaining a more fine-grained understanding of how well a classifier is performing rather than simply looking at total accuracy. The precision, sensitivity (recall), and F1-score are calculated using the following equations respectively.Table II illustrates the computed accuracy, recall, and F1-score values for the hockey fight and crowd violence datasets.
\begin{equation}
    Precision \hspace{.2em}=\hspace{.2em} \frac{True \hspace{.1em} Positive}{True \hspace{.1em}Positive + False \hspace{.1em}Positive}
\end{equation}
\begin{equation}
    Recall \hspace{.2em}=\hspace{.2em} \frac{True \hspace{.1em} Positive}{True \hspace{.1em}Positive + False \hspace{.1em}Negative}
\end{equation}
\begin{equation}
    F1 \hspace{.1em}score \hspace{.2em}=\hspace{.2em} 2 \times \hspace{.2em} \frac{Precision \times Recall}{Precision + Recall}
\end{equation}

\subsection{Comparative Discussion}
This subsection of the work draws a comparative analysis between the present study and previous state-of-the-art (SOTA) methods in the apposite task of violence detection from videos. The analysis has been considered for the hockey fight and violent crowd datasets. An elucidate comparison between the various SOTA approaches has been depicted in Table IV. Accuracies redeemed from both handcrafted-features as well as deep learning based approaches has been illustrated in the comparison. The proposed strategy explicitly achieves greater accuracy values in both benchmarking datasets, in comparison to prior systems in the literature that yield lower accuracy or are restricted to only one use case between person-to-person and crowd confrontations. Thus, the suggested technique is generalizable and applicable in a variety of circumstances while being highly accurate and computationally efficient.

\begin{table}[h]
	\centering
	\caption{Comparison of the proposed method with the state-of-the-art approaches}
	\label{table8}
	\begin{tabular}{|c|c|c|}
		\hline
	
		\textbf{Algorithm} &
		Hockey Fight&
		 Violent Crowd 
		\\

		\hline
		 ViF, OViF, AdaBoost and SVM \cite{GAO201637} & 
	     87.5\%  & 
		88\% 
		\tabularnewline
		\hline
		 Hough Forest and 2D CNN \cite{8375994}  & 
		 94.6\%  & 
		 $-$  
		\tabularnewline
		\hline
		 Three streams + LSTM \cite{10.1007/978-981-10-3002-4_43}& 
		 93.9\% & 
		$-$  
		\tabularnewline
		\hline
		Improved Fisher Vectors \cite{7738019} & 
		 93.7\% & 
		96.4\%  
		\tabularnewline
		\hline
		 3D Conv Net \cite{10.1007/978-3-319-14364-4_53} & 
		 91\% & 
		 $-$ 
		 \tabularnewline
		\hline
		CNN + LSTM \cite{8078468} & 
		 97\% & 
		94.57\%  
		\tabularnewline
		\hline

		\textbf{Augmentation, ViViT} & 
		 \textbf{97.14}\% & 
		 \textbf{98.46}\%
		\tabularnewline
		\hline

	\end{tabular}
\end{table}

\section{Conclusion and Future Works}

Detecting violence is critical for many applications, but its subjectivity makes it difficult for a generic model to be accurate. In the forefront of violence detection,
the present novel end-to-end framework has incorporated video vision transformers (ViViT) for efficient violent state estimation in video clips. Firstly, the video clips are casted into several frames and then, various image augmentation techniques are applied on the processed frames. The retrieved frames are passed on through a ViViT architecture which learns specific patterns from spatio-temporal information of the clips using the transformer encoder. The learnt patterns are utilised to categorise the film as violent or nonviolent. 
The suggested study  outperformed previous state-of-the-art methodologies in both person-to-person and crowd conflict datasets while being computationally efficient than CNN based approaches.

The potential downsides of the recommended architecture is that while the proposed model showed promising violent detection performance, it necessitates a large quantity of data with annotated scene conditions in order for the model to be trained better. So, generative adversarial networks (also known as GAN) can be introduced on the video frames to enlarge the training data. 

Furthermore, applying of different variants of the transformer model as discussed by Arnab et al. \cite{DBLP:journals/corr/abs-2103-15691} can also be considered for enhancing the performance of the model.

\bibliographystyle{IEEEtran}
\bibliography{references}

\begin{thebibliography}{10}
\providecommand{\url}[1]{#1}
\csname url@samestyle\endcsname
\providecommand{\newblock}{\relax}
\providecommand{\bibinfo}[2]{#2}
\providecommand{\BIBentrySTDinterwordspacing}{\spaceskip=0pt\relax}
\providecommand{\BIBentryALTinterwordstretchfactor}{4}
\providecommand{\BIBentryALTinterwordspacing}{\spaceskip=\fontdimen2\font plus
\BIBentryALTinterwordstretchfactor\fontdimen3\font minus
  \fontdimen4\font\relax}
\providecommand{\BIBforeignlanguage}[2]{{%
\expandafter\ifx\csname l@#1\endcsname\relax
\typeout{** WARNING: IEEEtran.bst: No hyphenation pattern has been}%
\typeout{** loaded for the language `#1'. Using the pattern for}%
\typeout{** the default language instead.}%
\else
\language=\csname l@#1\endcsname
\fi
#2}}
\providecommand{\BIBdecl}{\relax}
\BIBdecl

\bibitem{owidhomicides}
M.~Roser and H.~Ritchie, ``Homicides,'' \emph{Our World in Data}, 2013,
  https://ourworldindata.org/homicides.

\bibitem{2}
\BIBentryALTinterwordspacing
``Violent crime in the united states.'' [Online]. Available:
  \url{https://ucr.fbi.gov/crime-in-the-u.s/2019/crime-in-the-u.s.-2019/topic-pages/violent-crime}
\BIBentrySTDinterwordspacing

\bibitem{8375994}
I.~Serrano, O.~Deniz, J.~L. Espinosa-Aranda, and G.~Bueno, ``Fight recognition
  in video using hough forests and 2d convolutional neural network,''
  \emph{IEEE Transactions on Image Processing}, vol.~27, no.~10, pp.
  4787--4797, 2018.

\bibitem{19112472}
\BIBentryALTinterwordspacing
F.~U.~M. Ullah, A.~Ullah, K.~Muhammad, I.~U. Haq, and S.~W. Baik, ``Violence
  detection using spatiotemporal features with 3d convolutional neural
  network,'' \emph{Sensors}, vol.~19, no.~11, 2019. [Online]. Available:
  \url{https://www.mdpi.com/1424-8220/19/11/2472}
\BIBentrySTDinterwordspacing

\bibitem{8852616}
A.-M.~R. Abdali and R.~F. Al-Tuma, ``Robust real-time violence detection in
  video using cnn and lstm,'' in \emph{2019 2nd Scientific Conference of
  Computer Sciences (SCCS)}, 2019, pp. 104--108.

\bibitem{DBLP:journals/corr/abs-2010-11929}
\BIBentryALTinterwordspacing
A.~Dosovitskiy, L.~Beyer, A.~Kolesnikov, D.~Weissenborn, X.~Zhai,
  T.~Unterthiner, M.~Dehghani, M.~Minderer, G.~Heigold, S.~Gelly, J.~Uszkoreit,
  and N.~Houlsby, ``An image is worth 16x16 words: Transformers for image
  recognition at scale,'' \emph{CoRR}, vol. abs/2010.11929, 2020. [Online].
  Available: \url{https://arxiv.org/abs/2010.11929}
\BIBentrySTDinterwordspacing

\bibitem{DBLP:journals/corr/abs-2103-15691}
\BIBentryALTinterwordspacing
A.~Arnab, M.~Dehghani, G.~Heigold, C.~Sun, M.~Lucic, and C.~Schmid, ``Vivit:
  {A} video vision transformer,'' \emph{CoRR}, vol. abs/2103.15691, 2021.
  [Online]. Available: \url{https://arxiv.org/abs/2103.15691}
\BIBentrySTDinterwordspacing

\bibitem{https://doi.org/10.48550/arxiv.2209.01401}
\BIBentryALTinterwordspacing
G.~S. Krishna, K.~Supriya, J.~Vardhan, and M.~R. K, ``Vision transformers and
  yolov5 based driver drowsiness detection framework,'' 2022. [Online].
  Available: \url{https://arxiv.org/abs/2209.01401}
\BIBentrySTDinterwordspacing

\bibitem{1044748}
A.~Datta, M.~Shah, and N.~Da~Vitoria~Lobo, ``Person-on-person violence
  detection in video data,'' in \emph{2002 International Conference on Pattern
  Recognition}, vol.~1, 2002, pp. 433--438 vol.1.

\bibitem{1467545}
N.~Nguyen, D.~Phung, S.~Venkatesh, and H.~Bui, ``Learning and detecting
  activities from movement trajectories using the hierarchical hidden markov
  model,'' in \emph{2005 IEEE Computer Society Conference on Computer Vision
  and Pattern Recognition (CVPR'05)}, vol.~2, 2005, pp. 955--960 vol. 2.

\bibitem{5206569}
J.~Kim and K.~Grauman, ``Observe locally, infer globally: A space-time mrf for
  detecting abnormal activities with incremental updates,'' in \emph{2009 IEEE
  Conference on Computer Vision and Pattern Recognition}, 2009, pp. 2921--2928.

\bibitem{5539872}
V.~Mahadevan, W.~Li, V.~Bhalodia, and N.~Vasconcelos, ``Anomaly detection in
  crowded scenes,'' in \emph{2010 IEEE Computer Society Conference on Computer
  Vision and Pattern Recognition}, 2010, pp. 1975--1981.

\bibitem{10.5555/2044575.2044624}
E.~B. Nievas, O.~D. Suarez, G.~B. Garc\'{\i}a, and R.~Sukthankar, ``Violence
  detection in video using computer vision techniques,'' in \emph{Proceedings
  of the 14th International Conference on Computer Analysis of Images and
  Patterns - Volume Part II}, ser. CAIP'11.\hskip 1em plus 0.5em minus
  0.4em\relax Berlin, Heidelberg: Springer-Verlag, 2011, p. 332–339.

\bibitem{10.1007/978-981-10-3002-4_43}
Z.~Dong, J.~Qin, and Y.~Wang, ``Multi-stream deep networks for person to person
  violence detection in videos,'' in \emph{Pattern Recognition}, T.~Tan, X.~Li,
  X.~Chen, J.~Zhou, J.~Yang, and H.~Cheng, Eds.\hskip 1em plus 0.5em minus
  0.4em\relax Singapore: Springer Singapore, 2016, pp. 517--531.

\bibitem{SAMUELR2019191}
\BIBentryALTinterwordspacing
D.~J. {Samuel R.}, F.~E, G.~Manogaran, V.~G.N, T.~T, J.~S, and A.~A, ``Real
  time violence detection framework for football stadium comprising of big data
  analysis and deep learning through bidirectional lstm,'' \emph{Computer
  Networks}, vol. 151, pp. 191--200, 2019. [Online]. Available:
  \url{https://www.sciencedirect.com/science/article/pii/S1389128618308521}
\BIBentrySTDinterwordspacing

\bibitem{8078468}
S.~Sudhakaran and O.~Lanz, ``Learning to detect violent videos using
  convolutional long short-term memory,'' in \emph{2017 14th IEEE International
  Conference on Advanced Video and Signal Based Surveillance (AVSS)}, 2017, pp.
  1--6.

\bibitem{doi:10.1080/08839514.2020.1723876}
\BIBentryALTinterwordspacing
S.~Accattoli, P.~Sernani, N.~Falcionelli, D.~N. Mekuria, and A.~F. Dragoni,
  ``Violence detection in videos by combining 3d convolutional neural networks
  and support vector machines,'' \emph{Applied Artificial Intelligence},
  vol.~34, no.~4, pp. 329--344, 2020. [Online]. Available:
  \url{https://doi.org/10.1080/08839514.2020.1723876}
\BIBentrySTDinterwordspacing

\bibitem{DBLP:journals/corr/abs-2106-10270}
\BIBentryALTinterwordspacing
A.~Steiner, A.~Kolesnikov, X.~Zhai, R.~Wightman, J.~Uszkoreit, and L.~Beyer,
  ``How to train your vit? data, augmentation, and regularization in vision
  transformers,'' \emph{CoRR}, vol. abs/2106.10270, 2021. [Online]. Available:
  \url{https://arxiv.org/abs/2106.10270}
\BIBentrySTDinterwordspacing

\bibitem{GAO201637}
\BIBentryALTinterwordspacing
Y.~Gao, H.~Liu, X.~Sun, C.~Wang, and Y.~Liu, ``Violence detection using
  oriented violent flows,'' \emph{Image and Vision Computing}, vol. 48-49, pp.
  37--41, 2016. [Online]. Available:
  \url{https://www.sciencedirect.com/science/article/pii/S0262885616300063}
\BIBentrySTDinterwordspacing

\bibitem{7738019}
P.~Bilinski and F.~Bremond, ``Human violence recognition and detection in
  surveillance videos,'' in \emph{2016 13th IEEE International Conference on
  Advanced Video and Signal Based Surveillance (AVSS)}, 2016, pp. 30--36.

\bibitem{10.1007/978-3-319-14364-4_53}
C.~Ding, S.~Fan, M.~Zhu, W.~Feng, and B.~Jia, ``Violence detection in video by
  using 3d convolutional neural networks,'' in \emph{Advances in Visual
  Computing}, G.~Bebis, R.~Boyle, B.~Parvin, D.~Koracin, R.~McMahan, J.~Jerald,
  H.~Zhang, S.~M. Drucker, C.~Kambhamettu, M.~El~Choubassi, Z.~Deng, and
  M.~Carlson, Eds.\hskip 1em plus 0.5em minus 0.4em\relax Cham: Springer
  International Publishing, 2014, pp. 551--558.

\end{thebibliography}
 
\end{document}